# Addressing Data Distribution Shifts in Online Machine Learning Powered Smart City Applications Using Augmented Test-Time Adaptation

*Shawqi Al-Maliki*\*, *Graduate Student Member, IEEE*, *Faissal El Bouanani*†, *Senior Member, IEEE*, *Mohamed Abdallah*\*, *Senior Member, IEEE, Junaid Qadir*‡, *Senior Member, IEEE, Ala Al-Fuqaha*\*, *Senior Member, IEEE*

\* Information and Computing Technology (ICT) Division, College of Science and Engineering, Hamad Bin Khalifa University, Doha 34110, Qatar
† College of Engineering, Mohammed V University in Rabat, Morocco
‡ Department of Computer Science and Engineering, College of Engineering, Qatar University, Doha, Qatar

*Abstract*—Data distribution shift is a common problem in machine learning-powered smart city applications where the test data differs from the training data. Augmenting smart city applications with online machine learning models can handle this issue at test time, albeit with high cost and unreliable performance. To overcome this limitation, we propose to endow test-time adaptation with a systematic active fine-tuning (SAF) layer that is characterized by three key aspects: a *continuity* aspect that adapts to ever-present data distribution shifts; *intelligence* aspect that recognizes the importance of fine-tuning as a distribution-shift-aware process that occurs at the appropriate time to address the recently detected data distribution shifts; and *cost-effectiveness* aspect that involves budgeted human-machine collaboration to make relabeling cost-effective and practical for diverse smart city applications. Our empirical results show that our proposed approach outperforms the traditional test-time adaptation by a factor of two.

## I. INTRODUCTION

The performance of Machine Learning (ML) models in real-world applications relies heavily on the assumption that the data used during training and inference share the same distribution. However, this assumption is frequently broken in practice, leading to data distribution shifts that make ML models vulnerable. These shifts can occur naturally due to dynamic changes in the process or may be intentionally induced to compromise the model's performance. Regardless of their origin, data distribution shifts have been shown to significantly compromise the performance of ML models [1].

Traditional ML models are typically used in offline settings, where they are trained statically without any updates during testing. These models address data distribution shifts through periodic retraining using large datasets. However, this approach is inefficient and unreliable, and determining the optimal retraining time can be challenging. As a result, offline-learned ML models are unable to effectively handle distribution shifts, making them unsuitable for mission-critical applications.

Moreover, we live in an era of disruptive technologies where data is ever-changing. For example, IoT, smart cities, cyber-physical systems, metaverse, and industry 4.0 rely on ever-changing data, which necessitates addressing offline-learned ML models' drawbacks.

Online learning is a promising research direction that adopts a dynamic training approach to render ML models robust against distribution shifts. The use of dynamic models has been recommended by researchers such as Goodfellow [1] to make ML models robust to distribution shifts. In this work, *online learning* refers to any learning paradigm in which the model learns at test time. There are already some existing online learning paradigms such as online supervised learning, test-time adaptation, and online active learning, but these methods have inherent limitations that render them inappropriate for real-world smart city applications such as traffic management, public safety, medical diagnosis, autonomous driving, and visual defect detection. These limitations include:

1) *Online supervised learning* involves an expensive process that may not always be practical as true labels may not be available or may arrive with a delay [2].
2) *Test-time adaptation* (TTA) handles data distribution shifts at inference time by lightly changing the model parameters (less than 1%). That renders TTA insufficient at handling drastic data distribution shifts, where the model needs to be retrained and not only lightly updating 1% of its parameters.
3) *Online active learning*, does not incorporate the detection of drastic data distribution shift to the learning process. In addition, the retraining process requires the availability of the previous data, which is not always feasible due to privacy or commercial concerns.

In this work, we address natural data distribution shifts in an image classification task as a representative proxy for handling different sources of data distribution shifts. We seek to establish a realistic online learning setting that overcomes the limitations of existing online learning paradigms. Fig. 1 provides a taxonomy that illustrates how our work relates to existing online learning paradigms.

Typical TTA has an attractive online setting amongst the existing online learning paradigms. The most appealing characteristic of this variant of online learning is relaxing the necessity of having the ground truth labels, which fits nicely in vision tasks where getting the ground truth timely and cost-effectively is a big challenge. However, TTA techniques do not sufficiently adapt to highly changing environments where the model needs to be retrained and not only lightly updating 1% of its parameters. That renders these models inappropriate in real-world smart city applications. For that, we propose augmenting TTA with Systematic Active Fine-tuning (SAF). SAF is a metric-based fine-tuning process that, reliably and cost-effectively, detects and corrects drastic data distribution shifts in online machine learning powered smart city systems (details in Section III). It incorporates continuity, intelligence, and cost-effectiveness aspects: *continuity* indicates that models confront the ever-ending data distribution shifts at the test time, *intelligence* implies that fine-tuning is a distribution-shift-aware process that occurs at the appropriate time to address the recently detected data distribution shifts, and *cost-effectiveness* aspect indicates employing budgeted human-machine collaboration for the relabeling to be cost-effective and practical for diverse smart city applications. An abstract illustration of our scheme is presented in Fig. 2, while a more detailed description of SAF follows in Section III-B).

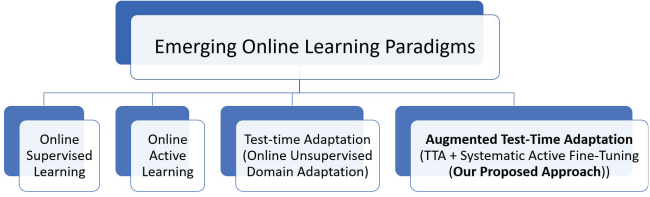

Fig. 1: A taxonomy contextualizing our work among other emerging online learning paradigms that can be used in smart city applications.

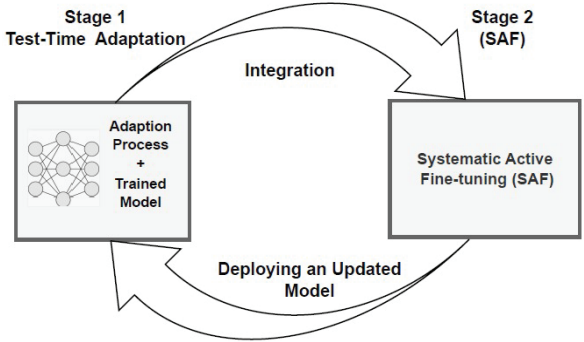

Fig. 2: Abstract view of the proposed *augmented TTA* (TTA + SAF) as a realistic online setting to address significant distribution shifts encountered by critical-mission smart city applications.

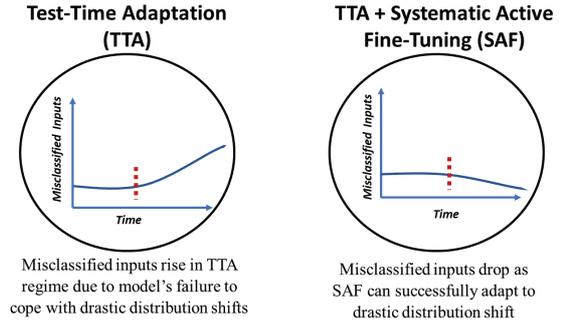

Fig. 3: An example demonstrating enhanced model performance in scenarios with drastic data distribution shifts.

Fig. 3 provides an example that shows how our work enhances model's performance in an urban environment with drastic data distribution shifts. It illustrates that misclassified inputs are minimized when our proposed approach is employed.

The salient contributions of this work are as follows.
- We highlight the weakness of test-time adaptation (TTA)

systems in highly changing urban environments.
- We propose a generic online scheme that integrates TTA with SAF to handle drastic distribution shifts in a fast-changing environment.
- We perform extensive simulations to validate the proposed scheme.
- We highlight promising future research directions towards realizing fully online learning in practical settings.

The rest of this work is organized as follows. We discuss related works and identify research gaps therein in Section II. We then delineate the details of our proposed scheme in Section III. Subsequently, we present experimental results that validate our proposed approach in Section IV. After that, we illustrate potential smart city applications in Section V. Finally, we highlight promising research directions in Section VI.

## II. RELATED WORK

In the following subsections, we review relevant online learning paradigms and highlight how our proposed approach differs from these approaches.

### A. Online Supervised Learning

Online supervised learning performs training by incrementally and continuously updating the model on each instance in the data stream. It is robust to data distribution shifts as it handles the worst-case data distribution shift where every single input is a concern. However, being expensive and functionally infeasible for handling tasks such as image classification, makes it a restricted learning paradigm. It fits well with a few ML tasks in which the true labels come automatically after some time (e.g., delivery time estimation). The work of Yang and Shami [3] is an example of the works in this research direction, where tabular data is used, and the true labels come automatically after some time. Hoi et al. [4] present further works with online supervised setting.

### B. TTA Paradigm

TTA is an emerging online learning paradigm [6], [8], [12] that utilizes only unlabeled data to continuously adapt the model. It fits application domains where the ground truth is financially or operationally infeasible. Examples of the adaptation techniques used in this research direction are batch normalization [7], entropy minimization [8], or self-learning [5], [6]. The adaptation techniques incorporated in TTA may help in maintaining the trained model without retraining for a while. Specifically, when the environment is not changing drastically.

TABLE I: SUMMARY of RELATED ONLINE LEARNING Paradigms

| Related Work | Learning Paradigm | Robustness Against Natural Dist. Shifts | Training Efficiency (Required Labels) | | Suitability for Image Classification Task? | Practical for Smart City Application? |
|---|---|---|---|---|---|---|
| | | | Required Labels | Training Mode For Handling Dist. Shifts | | |
| [3] [4] | Online Supervised Learning | Yes | Full Labels | Continual Training | No | With Costly Monitoring |
| [5] [6] [7] [8] | Test-Time Adaptation (TTA) | Partially | No Labels | Continual Adaptation | Yes | Insufficient Robustness |
| [9] [10] | Online Active Learning | No | Limited Labels | N/A | Yes | No |
| [11] | Transfer Learning By Fine-Tuning | No | Limited Labels | N/A | Yes | No |
| **Our work** | **SAF with TTA** | **Yes** | **Limited Labels** | **Continual Fine-tuning** | **Yes** | **With Sufficient Robustness** |

### C. Online Active Learning

In this paradigm, the learner is assumed to have the training data along with the whole set of the selected and relabeled data to perform retraining from scratch. It does not consider the concern of data distribution shift. In particular, it doesn't detect and address multiple distribution shifts. [9], [10] are previous studies on online active learning.

### D. Transfer learning by Fine-Tuning

Transfer learning by fine-tuning is well-motivated in situations with abundant training data in a source domain and less data on the target domain. Typical fine-tuning [11] involves a one-time adaptation of one domain to another. Also, it happens in an offline setting and does not involve a systematic process that detects and corrects the change in data distribution. In addition, although a typical fine-tuning occurs by utilizing a small set of inputs, it does not include an active selection process to choose the most informative data for the fine-tuning process. Thus, it is not practical for real-world smart city applications where data distribution shift is a concern.

Existing online learning paradigms do not account for real-world scenarios where a certain level of robustness against data distribution shifts is required. Among existing online learning paradigms, typical TTA is attractive yet restrictive in real-world smart city applications where encountering continual drastic data distribution shifts is typical. Thus, our proposed approach is based on augmenting it with SAF to form a practical online learning paradigm.

### E. Our Proposed Online Learning Approach

TTA lightly changes the model parameters (less than 1%) [8] as detailed in Section III-A. This feature can be seen as an advantage as it reduces computational overhead, which renders it an effective online technique. On the other hand, this feature makes TTA insufficient at handling drastic changes in input distribution. That is, there are real-world situations where typical TTA is not enough, and the model needs to be retrained and not only lightly updating 1% of its parameters. Fine-tuning the model incorporated in TTA, and not only lightly adapting it, is the proper solution to address drastic changes in data distribution. However, traditional fine-tuning is periodic, which means it can happen unnecessarily, i.e., wasting time and resources. Furthermore, traditional fine-tuning utilizes the typical full labeling (i.e., annotating all available data), which is expensive. In addition, fine-tuning adapts a pre-trained model to one data distribution shift (i.e., a new relevant task). In real-world smart city applications, there can be many types of distribution shifts, and each needs to be addressed differently. On the other hand, efficient techniques that identify these types of shifts cost-effectively are not incorporated into traditional fine-tuning. Thus, there is a necessity for systematic active fine-tuning that occurs whenever a drastic change in data distribution is identified. We propose integrating TTA with systematic active fine-tuning to form a practical online setting for handling drastic data distribution shifts in smart city applications.

Table I summarizes the merits and limitations of the learning paradigms related to our proposed approach. It demonstrates that our proposed approach addresses the weaknesses of related learning paradigms.

## III. INTEGRATING TTA WITH SYSTEMATIC ACTIVE FINE-TUNING

We present a more detailed view of *SAF* in Fig. 4 to expand upon the abstract view presented earlier in Fig. 2. *SAF* starts by selecting the most informative inputs, moving them to human experts for validation & relabeling, and then extracting the distribution-shift-related information (i.e., the types and levels of changes in data distribution). After detecting that an adaptation insufficiency, a fine-tuning process can be triggered either automatically or by a human expert. Fine-tuning the trained model incorporated in the TTA stage is done utilizing the subset of relabeled data that is relevant to the detected type of data distribution shift.

### A. TTA Stage

The TTA stage receives a stream of images, processes them, and generates corresponding output confidences as soft predictions. It aims to handle distribution shifts at inference time without the help of labeled data. The TTA paradigm is a special type of unsupervised domain adaptation that maps a source a trained model to a target test-time self-adapted model. This paradigm comprises the trained model and an adaptation technique. A self-adaptation technique such as entropy minimization [8] keeps updating very few batch normalization relevant parameters (less than 1%) of the trained model without updating the weights of the model. At test time, a self-adapted model utilizes unsupervised learning approaches to capture the potential out-of-distribution inputs after their first-time prediction. It updates a few parameters of the trained model directly without a retraining process. Thus, the model adapts and correctly predicts, to some extent, the subsequent out-of-distribution inputs.

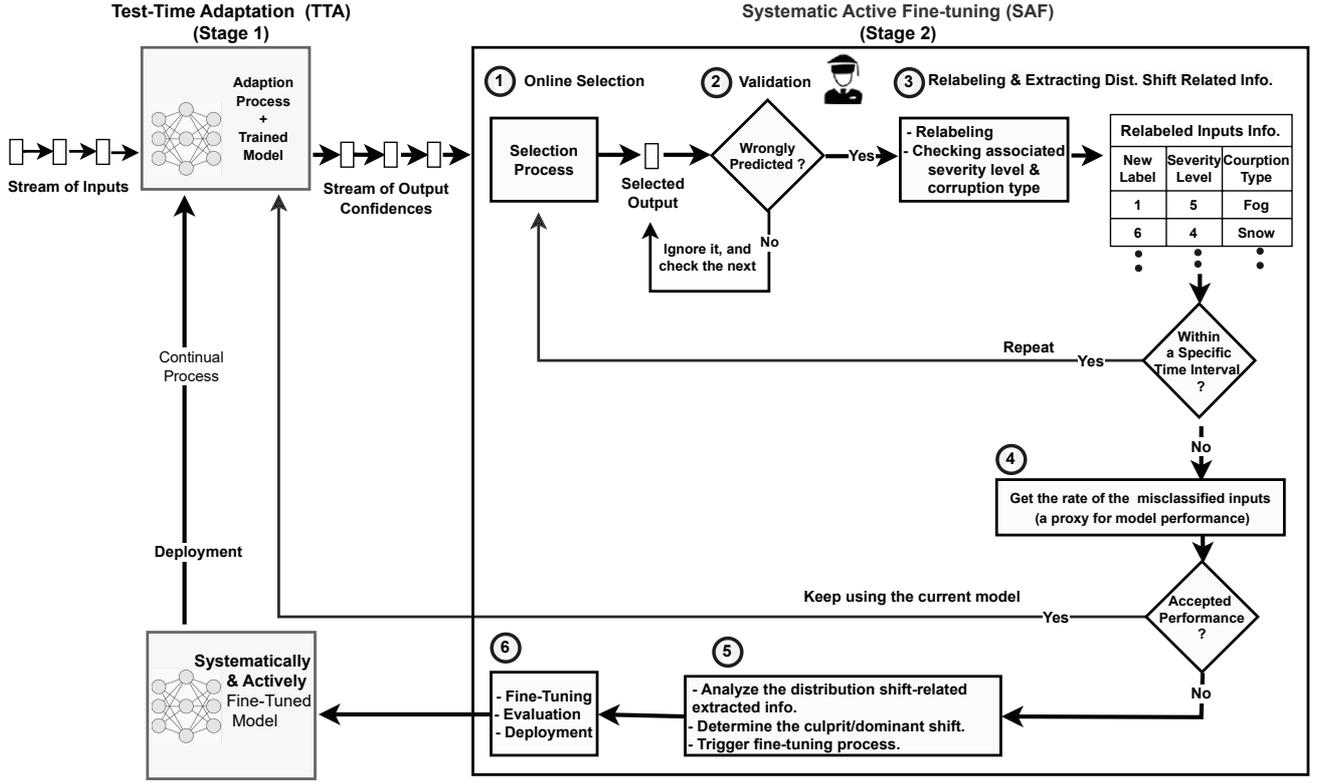

Fig. 4: Detailed Proposed Approach, where TTA (Stage 1) complements SAF (Stage 2) forming an *augmented TTA* paradigm that can be utilized by mission-critical smart city applications to address drastic data distribution shifts at inference time. SAF strives to generate a fine-tuned model that replaces the trained model when necessary.

This TTA approach does not sufficiently adapt to highly changing environments such as smart cities. Thus, it is inadequate for handling smart city applications. Future AI solutions should consider complementing such learning approach with other robust yet cost-effective techniques. To this end, we propose integrating it with SAF.

### B. Systematic Active Fine-tuning (SAF) Stage

To maintain the required performance of TTA in mission-critical smart city applications, we should have an ongoing monitoring and evaluation process to know how well they are working. On the other hand, measuring model performance requires having the true labels of the predicted inputs. For vision tasks such as image classification, true labels cannot be obtained automatically. Instead, human experts should be involved in validating the predicted inputs and relabeling all wrongly predicted ones. Then the retraining process starts. Often, it is a tedious, ineffective, and financially infeasible task. Thus, we propose enhancing the monitoring process by incorporating SAF. SAF is a metric-based fine-tuning process that occurs as part of monitoring and evaluating the performance of the deployed smart city systems. It is characterized by continuity, intelligence, and cost-effectiveness. *Continuity* implies that fine-tuning is an ongoing process that happens whenever a data distribution shift is detected. *Intelligence* has two aspects (the *When* and *How*). In the *When* aspect, *intelligence* means that fine-tuning is not performed blindly but intelligently at the time a data distribution shift is detected. Thus, it avoids unnecessary fine-tuning. The *How* aspect of *intelligence* means that the fine-tuning doesn't utilize the whole relabeled data (curated data), but rather it intelligently utilizes the subset of the relabeled data that is most relevant to the latest detected data distribution shift. The *cost-effectiveness* implies that the detection of data distribution shift is based on an opportunistic selection of the most informative inputs to be validated and relabeled by human experts under a budget-constrained situation. The steps that comprise SAF are described next.

*1) Online Selection:* Utilizing windowing-based online selection techniques, the most informative images (i.e., images with the least confidence) are selected and pushed to the validation process (Step 1 in Fig. 4). In online selection settings where the data stream is never-ending and selected images cannot be revoked, we have to decide on the fly which images to choose for validation and which ones to ignore. We urge considering intelligent and efficient online selection techniques that contribute to the efficient utilization of inspection, analysis, and fine-tuning resources. For example, a windowing-based online selection technique such as Opportunistic Selection and Relabeling Algorithm (OSRA) [13] can be utilized as done in this work.

*2) Validation:* For the image classification task, the human expert (assumed to be reliable) can judge whether the images that come from Stage 1 have correct predictions by comparing the actual representation of the image with the predicted label (Step 2 in Fig. 4). If there is a mismatch, the associated images are moved to the next step (relabeling and extracting the distribution shift-related information).

*3) Relabeling and Extracting Distribution Shift-Related Information:* In this step, the human expert receives the wrongly predicted images (the mismatched ones) and starts processing them. In particular, he relabels an image and extracts the distribution shift-related information such as the degree and the type of the change (also called the severity level and the corruption type) by comparing them with the sample images in the benchmarked dataset (RobustBench [14]). Then, the extracted information is appended as a new record in the curated data. A curated data is a table that tracks the distribution shift-related information of the relabeled inputs (Steps 3 in Fig. 4). The process of validation, relabeling, and extracting the distribution shift information iterates in a timely fashion (e.g., day).

*4) Measuring Model Performance:* At the end of a specific time interval, the curated data is analyzed to check the rate of the wrongly predicted images to be used as a proxy for the prediction process performance (Step 4 in Fig. 4). If the performance of the model is above a specific threshold (application-specific value), the current model is kept. Consequently, unnecessary fine-tuning is avoided. It prevents unneeded usage of computer resources (computational-wise optimization). On the other hand, if the model's performance is found below a specific threshold, it indicates the right time for fine-tuning (the "when" aspect of intelligent fine-tuning).

*5) Analyzing and Triggering Fine-tuning:* Then, the curated data is analyzed further to check the culprits. Culprits are the dominant corruption types, i.e., data distribution shift types, with higher severity levels, i.e., the degree of data distribution shift. They contribute to deteriorating the prediction process of the model. Thus, the corresponding relabeled images of those culprits are extracted from the curated data and utilized for fine-tuning the deteriorated model.

Lastly, the model gets fine-tuned, evaluated, and re-deployed (Step 6 in Fig. 4).

We note that relabeling process has a dual purpose—detection and correction of data distribution shifts. The relabeling process contributes to measuring the rate of the misclassified inputs and accordingly indicates the existence of the data distribution shift i.e., deterioration. Also, the relevant relabeled inputs are utilized while performing the fine-tuning process i.e., correction.

At the end of a time unit, the whole set of the relabeled data is processed to uncover the potential existence of data distribution shifts. Specifically, the samples with the same type of data distribution shifts are aggregated to form groups of natural data distribution shift types (e.g., snow, fog, frost). Then, they are ordered to recognize the group with the highest number of samples (i.e., the dominant group). This is interpreted as data distribution shift detection. The next is the correction action, which is simply utilizing the samples of the dominant group in a fine-tuning process that results in an adapted model that performs well when it encounters data with a similar distribution shift. Triggering the fine-tuning process is application dependent. It can be automated or it can be triggered by a human.

## IV. Experiments and Results

### A. Experimental Setup

To show the effectiveness of our proposed approach in addressing the drastic data distribution shift of smart city applications, we simulate several data distribution shifts in a fast-changing environment utilizing RobustBench [14] which is a standard benchmark for evaluating model robustness against natural and adversarial distribution shifts. We utilize CIFAR-10-c as a representative dataset because of its popularity as a benchmark for evaluating natural distribution shifts [14]. We used CIFAR-10-c as a proxy for smart city datasets and applications to be able to compare our work with the existing works because CIFAR-10-C is heavily used in the literature that handles data distribution shifts for image classification tasks. It is the most common benchmark that facilitates simulating different types of natural distribution shifts (e.g., fog, snow, frost). Also, it facilitates simulating various levels (e.g., 1, 2, 3) of natural distribution shifts such that higher levels mean a more drastic change in data distribution.

CIFAR-10-C is the corrupted version of the standard CIFAR-10 dataset. We also utilize test-time adaptation by entropy minimization [8] that is based on a WRN-28-10 trained model, which is the architecture of a standard model in the RobustBench benchmark [14]. A simulated input image in CIFAR-10-C is associated with a corruption type and a severity level of the corruption, i.e., the degree of the detected data distribution shift. We simulate five environmental corruption types (fog, snow, frost, contrast, brightness) and five severity levels (1, 2, 3, 4, 5). Higher severity levels imply higher distribution shifts.

We simulate two scenarios utilizing the severity levels and the corruption types of the benchmark. Each scenario represents a potential environmental change that a model encounters in smart city applications.

A scenario is a stream of 1680 images, which is the total images of seven consecutive time intervals (e.g., seven days (a week)). That is, a stream of 240 images in a time interval of a time scale of one day. In the conducted experiments, we allocated 10% budget units for the human intervention process. Allocating 10% budget units means allocating budget units that are sufficient for handling 10% of the elements in the stream. We assume that one allocated budget unit represents the required human effort for handling one element. That means that the allocated budget for handling 24 elements (10% of the stream) is 24 budget units. Allocating 10% budget units is just an example. The allocated budget can be more or less based on the available budget for relabeling and the required enhancement for the application. In other words, it is budget- and -application dependent. OSRA [13]) shows further examples of allocated budgets of 5% and 15%. In this work, we adopted the online selection approach (OSRA) for data sampling. All results are based on the average of 100 experimental replications.

*1) Performance Metric:* In our experiments, we utilized *the rate of misclassified inputs* as a proxy for measuring the performance of the deployed online ML-powered smart city systems. It is a proxy for the model performance. Standard metrics, such as accuracy, F1-Score, precision, and recall, are usually used for evaluating models in offline settings during the development phase, where all the ground truth labels are available. However, for evaluating models in online settings, especially image-based ML tasks, having all the ground truth labels of the predicted samples is challenging. For that, proxy metrics are a common practice.

In our work, we use "model performance" and "robustness" interchangeably to refer to model robustness against natural data distribution shifts.

*2) Research Questions:* We perform extensive experiments to answer the following questions:

- How does our proposed solution handle drastic data distribution shifts that occur in a constantly changing environment?
- How do windowing-based online selection techniques contribute to the performance of the proposed solution?
- How is the impact of *SAF* on *TTA* enhancement?

### B. Experimental Scenarios

The first simulated scenario depicted in Fig. 5a refers to a situation where no data distribution shifts are detected in the considered time intervals for evaluating the performance of the deployed smart city system. Thus, fine-tuning is not triggered. This scenario demonstrates the best-case situation from the data distribution shift perspective (distribution-shift wise) and the worst-case situation from the budget-utilization perspective (budget-utilization wise). The use of the limited budget in this scenario is justified as a proactive step in real-world application domains where the uncertainty of experiencing distribution shifts is remarkably high.

The second scenario illustrated in Fig. 5b illustrates a potential situation with a highly-changing environment. The scenario is based on encountering a new type of drastic distribution shift periodically after two subsequent time intervals. The intelligent fine-tunings have mitigated the risk of model performance deterioration from a complete deterioration in the considered time range to a partial deterioration. Model performance deterioration occurs as a result of data distribution shift. *SAF* detects the data distribution shift as soon as it is encountered in a time interval. Then, it intelligently triggers fine-tuning to avoid model deterioration in subsequent time intervals. This scenario represents the best-case situation (budget-utilization wise) and the worst-case situation (distribution-shift wise).

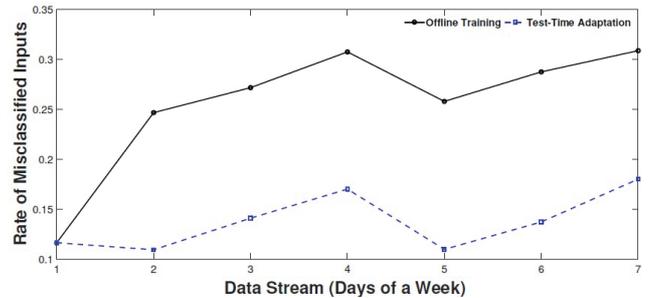

(a) *Scenario 1*: *No data distribution shifts detected in the considered time intervals*. Fine-tuning is not triggered. This scenario demonstrates the best-case situation distribution-shift-wise, and the worst-case situation budget-utilization-wise. The use of the limited budget in this scenario is justified as a proactive step in real-world application domains where distribution shifts are highly uncertain.

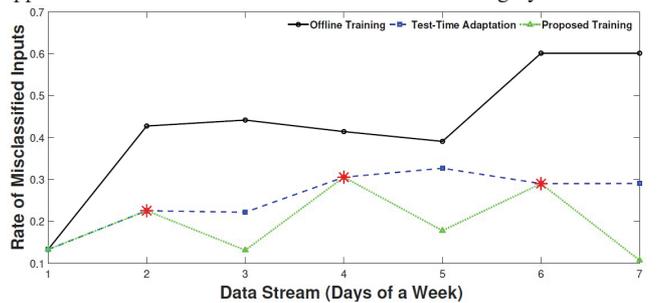

(b) *Scenario 2*: *Frequent time fine-tunings per considered time interval. This scenario represents the best-case situation (budget-utilization wise) and the worst-case situation (distribution-shift wise).* A new type of drastic distribution shift in each two subsequent time intervals. The intelligent fine-tunings have mitigated the risk of model performance deterioration from a complete deterioration in the considered time-intervals.

Fig. 5: Performance evaluation of the proposed approach compared to offline training and TTA.

In all investigated scenarios, the results of our proposed learning paradigm outperformed offline learning (on average) by a factor of 3 and outperformed TTA by a factor of 2.

### C. Windowing-based vs Random-based Selection Strategies

In this section, we aim to show the effectiveness of utilizing windowing-based strategies for selecting the most informative inputs where data distribution shift is a concern. As detailed in Section III-B1, we adopt the use of a window-based selection strategy in our work. We use the second simulated scenario, as an example, to demonstrate the success rate of selecting the most informative inputs when using both the random-based and windowing-based selection strategies. Fig. 6 clearly illustrates that windowing-based selection strategies outperform random-based selection strategies by a significant margin. Thus, we recommend using windowing-based selection techniques when considering *SAF* for handling data distribution shifts.

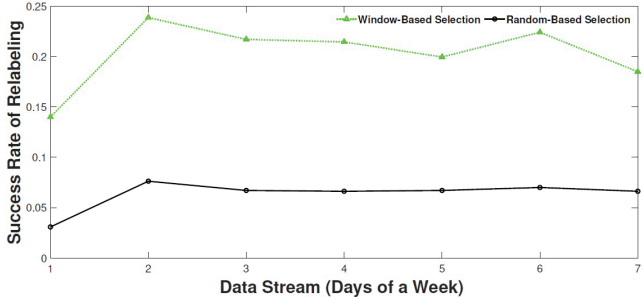

Fig. 6: The rate of the successful selection of the most informative inputs when we utilize windowing vs. random selection algorithms (the higher the better). Windowing outperforms random selection by a significant margin. *Thus, we urge utilizing windowing-based selection techniques when considering (SAF).*

### D. Lessons Learned

- Without utilizing SAF where a drastic data distribution shift is a concern, there is a high potential that a deployed smart city application suffers from longer-time performance deterioration.
- Complementing TTA with *SAF* renders TTA a practical real-world online learning paradigm.
- SAF is a perpetual process which means that the corresponding cost of human-machine interaction is continuous. For that, our work is notable when the potential of the occurrence of multiple data distribution shifts is high such as in high-stakes smart city applications.
- For the image classification task, the reliable detection and correction of data distribution shifts require expensive human-in-the-loop (HitL) validation and tweaking. Our work optimizes the HitL process so it becomes attractive to smart city system operators to utilize it to gain more reliability and save more budget.
- Detection of data distribution shifts requires investigating at least one-time interval. So, there will always be unavoidable potential exposure to unhandled data distribution shift equals one-time interval of a specific time scale. Thus, the shorter the time interval, the better the handling of data distribution shifts.

## V. POTENTIAL SMART CITY APPLICATIONS

Our proposed work is generic and can be used in various smart city applications.

### A. Traffic Management

Handling the traffic flow timely and efficiently is vital for the well-being and safety of urban residents. Incorporating online ML models in traffic management systems can help with addressing traffic issues effectively. Adopting our proposed *augmented TTA* in the traffic management solutions can robustify the timely identification of congestion issues and any anomalous traffic patterns, such as road closures and construction zones, to trigger proper actions including rerouting the traffic and adjusting traffic signal timings.

### B. Public Safety

Incorporating our proposed *augmented TTA* into the surveillance systems in public urban environments can help to robustify the recognition of anomalous behavior of city residents and visitors. The robust identification of such safety-critical patterns can be utilized as warnings submitted to local authorities to react early and save resident lives in urban environments.

### C. Medical Dignosis

The change in the distribution of health data is constantly changing. Thus, traditional online machine learning models become less accurate over time. a solution incorporated with intelligent fine-tuning based on new data patterns is essential. Our proposed *augmented TTA* can serve as a dependable and adaptable incorporated approach in the medical diagnosis systems to optimize the robustness of those systems so that they can identify emerging disease patterns efficiently. For that, urban systems adopting *augmented TTA* can provide more targeted interventions for at-risk patients.

### D. Autonomous Driving

Autonomous vehicles generate a massive amount of data from sensors and cameras while they navigate urban areas. However, the captured inputs constantly change, which negatively impacts the accuracy of ML models incorporated in autonomous driving management systems. The deteriorated performance of those models puts the lives of passengers and pedestrians at risk. Our proposed *augmented TTA* can contribute efficiently to optimizing the models' predictive power and thus mitigates the risks by utilizing newly collected, yet with distribution-shift, data during real-world driving to systematically and actively fine-tune the initially trained model incorporated in the prediction system.

### E. Smart Manufacturing

Our proposed online learning paradigm can be used as part of cyber-physical systems to control the quality of products efficiently. For example, mobile device manufacturers can use it as *ML-based visual defect detection* solution to optimize the efficiency of identifying and fixing defective devices on the production line. When a device reaches the quality checking stage, our system scans it to predict its status and assigns (on the fly) a label of defective, perfect, or suspected based on the system's degree of prediction confidence. The suspected devices are moved to human experts to validate the status and relabel them so they can be moved confidently to defective or perfect device queues. Without this smart intervention of a human expert, a customer may get a defective device. After that, the images of relabeled mobiles are used for fine-tuning the model to enhance its predictive power for future use.

## VI. FUTURE DIRECTIONS

In this section, we highlight the potential works that might contribute to moving online learning paradigms towards realistic online settings that can ideally serve to enhance the robustness of smart city systems.

### A. Potential Impact of our Work on ML Models Security Concerns

*1) Test-Time Security Concerns:* Traditional ML-powered smart city systems are susceptible to various adversarial attacks. For example, they are brittle to adversarial examples. We highly encourage the smart city and online learning research communities to investigate the impact of our proposed scheme on the adversarial robustness of smart city applications. We believe that our generic scheme contributes to adversarial robustness because it makes the deployed smart city system a moving target. That makes the process of generating adversarial examples more difficult. If an adversary managed to craft an adversarial example during a specific version of the model. In that case, this recently generated attack most probably will not be valid for attacking the model because it will adapt to a different version giving it the defender's last-move advantage.

*2) Training-Time Security Concerns:* A common security concern in ML-powered smart city applications is training-time poisoning attacks. They occur when an adversary manages to access and manipulate the data used for training a model. Since our proposed scheme enhances ML model robustness by utilizing unlabeled data (Stage 1) and limited labeled data (Stage 2), that means that our proposed scheme minimizes the model's exposure to more labeled data, which is the source of poisoning attacks. That is because our proposed scheme avoids training ML models from scratch using large labeled data. We believe that studying the impact of our generic scheme on enhancing ML-powered smart city applications' robustness against training-time poisoning attacks is a promising future direction.

### B. Dynamic Windowing for Effective Online Selection

We believe that a sliding window that shrinks and expands dynamically, based on the success frequency of the previous selections, fits well with the online setting of the proposed scheme. Incorporating a dynamic sliding window with our proposed scheme will enhance the scheme's robustness. Thus, we encourage the online learning research community to investigate the impact of utilizing dynamic windowing in the online selection process on the performance of the proposed scheme.

## VII. CONCLUSION

This work arms the researchers and practitioners of online learning and smart city communities with a generic approach that renders the typical test-time adaptation (TTA) approach a practical solution in mission-critical smart city applications. TTA is an emerging cost-effective research direction yet does not sufficiently handle drastic data distribution shifts. We propose an augmented variation of TTA that outperforms typical TTA. Specifically, it augments traditional TTA with a systematic active fine-tuning (SAF) layer to render it appropriate for handling drastic data distribution shifts in smart city applications. The added layer (SAF) has continuity, intelligence, and cost-effectiveness aspects. Continuity implies confronting the ever-ending data distribution shifts. The intelligence aspect means that the fine-tuning happens at the time a distribution shift is detected and utilizes the most relevant subset of the relabeled data. The cost-effectiveness aspect implies that budgeted human-AI collaboration is considered so relabeling becomes feasible for various smart city applications. This work opens the doors for the smart city research community to conduct further research toward more robust and efficient online ML-powered smart city systems. The community can use the proposed generic approach as a base to explore promising research directions, such as robustifying ML security concerns at training and test time and adopting dynamic windowing for effective relabeling.

## VIII. ACKNOWLEDGMENT

This publication was made possible by NPRP grant # [13S-0206-200273] from the Qatar National Research Fund (a member of Qatar Foundation). The statements made herein are solely the responsibility of the authors.

## REFERENCES

[1] I. Goodfellow, "A research agenda: Dynamic models to defend against correlated attacks," *In Proc. of ICLR, pages 1–9, New Orleans, LA, USA*, May 2019.

[2] H. M. Gomes, M. Grzenda, R. Mello, J. Read, M. H. Le Nguyen, and A. Bifet, "A survey on semi-supervised learning for delayed partially labelled data streams," *ACM Computing Surveys (CSUR)*, 2022.

[3] L. Yang and A. Shami, "A lightweight concept drift detection and adaptation framework for iot data streams," *IEEE Internet of Things Magazine*, vol. 4, no. 2, pp. 96–101, 2021.

[4] S. C. Hoi, D. Sahoo, J. Lu, and P. Zhao, "Online learning: A comprehensive survey," *Neurocomputing*, vol. 459, pp. 249–289, 2021.

[5] E. Rusak, S. Schneider, P. Gehler, O. Bringmann, W. Brendel, and M. Bethge, "Adapting imagenet-scale models to complex distribution shifts with self-learning," *arXiv preprint arXiv:2104.12928*, 2021.

[6] J. Liang, D. Hu, and J. Feng, "Do we really need to access the source data? source hypothesis transfer for unsupervised domain adaptation," in *International Conference on Machine Learning*. PMLR, 2020, pp. 6028–6039.

[7] S. Schneider, E. Rusak, L. Eck, O. Bringmann, W. Brendel, and M. Bethge, "Improving robustness against common corruptions by covariate shift adaptation," *Advances in Neural Information Processing Systems*, vol. 33, pp. 11 539–11 551, 2020.

[8] D. Wang, E. Shelhamer, S. Liu, B. Olshausen, and T. Darrell, "Tent: Fully test-time adaptation by entropy minimization," in *International Conference on Learning Representations*, 2021. [Online]. Available: https://openreview.net/forum?id=uXl3bZLkr3c

[9] J. Lu, P. Zhao, and S. C. Hoi, "Online passive-aggressive active learning," *Machine Learning*, vol. 103, no. 2, pp. 141–183, 2016.

[10] S. Hao, J. Lu, P. Zhao, C. Zhang, S. C. Hoi, and C. Miao, "Second-order online active learning and its applications," *IEEE Transactions on Knowledge and Data Engineering*, vol. 30, no. 7, pp. 1338–1351, 2017.

[11] J. Yosinski, J. Clune, Y. Bengio, and H. Lipson, "How transferable are features in deep neural networks?" *Advances in neural information processing systems*, vol. 27, 2014.

[12] Y. Sun, X. Wang, Z. Liu, J. Miller, A. Efros, and M. Hardt, "Test-time training with self-supervision for generalization under distribution shifts," in *International Conference on Machine Learning*. PMLR, 2020, pp. 9229–9248.

[13] S. Al-Maliki, F. El Bouanani, K. Ahmad, M. Abdallah, D. Hoang, D. Niyato, and A. Al-Fuqaha, "Toward improved reliability of deep learning based systems through online relabeling of potential adversarial attacks," 2022.

[14] F. Croce, M. Andriushchenko, V. Sehwag, E. Debenedetti, N. Flammarion, M. Chiang, P. Mittal, and M. Hein, "Robustbench: a standardized adversarial robustness benchmark," *arXiv preprint arXiv:2010.09670*, 2020.